\documentclass[letterpaper, 10 pt, conference]{ieeeconf}

\IEEEoverridecommandlockouts
\usepackage{graphicx}
\usepackage{amsmath,amssymb}
\usepackage{booktabs}
\usepackage{multirow}
\usepackage{cite}
\usepackage{url}
\usepackage{xcolor}
\usepackage{placeins}
\usepackage{float}
\usepackage{capt-of}
\usepackage{balance}

\title{\LARGE \bf TemporalFlow-VLA: Learning Physically Grounded Execution History for Long-Horizon Robot Manipulation}
\author{Jiarui Yang$^{1}$, Yehao Lu$^{2}$, Yuning Su$^{3}$, Yu Zhong$^{4}$, Yufeng Xie$^{4}$, Yazhou Zhang$^{4}$,\\
Haiyu Lan$^{4}$, Kaixiang Lu$^{4}$, Peiwen Lin$^{4}$, Chuang Wang$^{4}$, Junwei Liang$^{1,*}$, and Enyu Li$^{4,*}$%
\thanks{$^{1}$The Hong Kong University of Science and Technology (Guangzhou), Guangzhou, China.}%
\thanks{$^{2}$Zhejiang University, Hangzhou, China.}%
\thanks{$^{3}$Simon Fraser University, Burnaby, BC, Canada.}%
\thanks{$^{4}$AgiBot, Shanghai, China.}%
\thanks{$^{*}$Corresponding authors.}%
}

\IEEEaftertitletext{%
\vspace{0.2em}
\begin{minipage}{\textwidth}
    \centering
    \includegraphics[width=\textwidth]{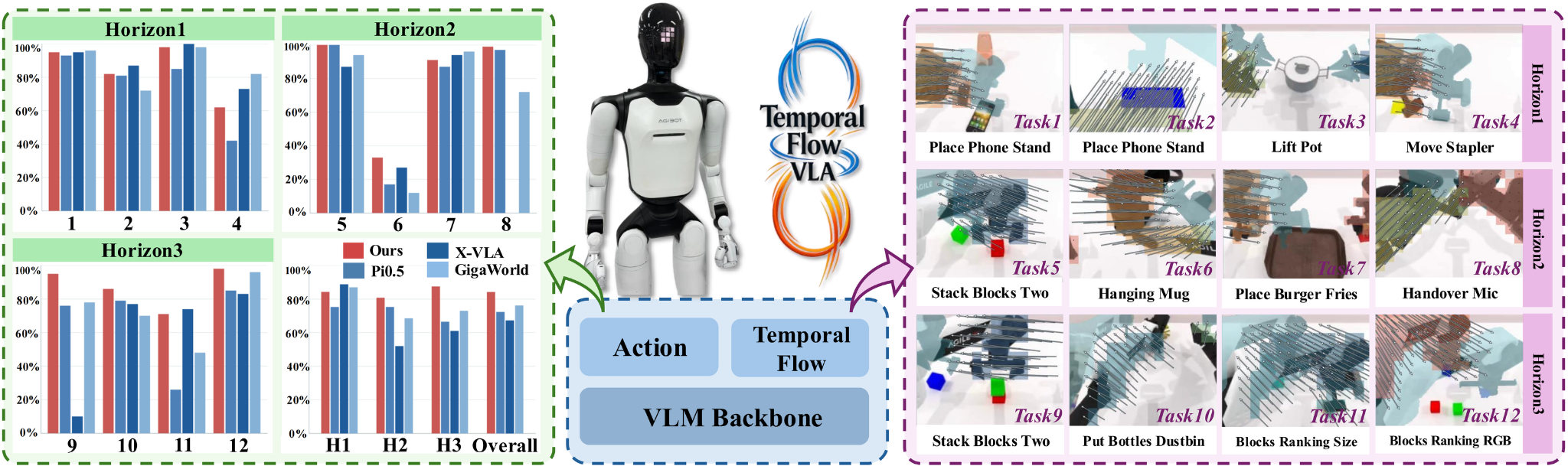}
    \captionof{figure}{Overview of TemporalFlow-VLA. \textbf{Left:} randomized RoboTwin 2.0 success rates over all 12 tasks and the H1/H2/H3/Overall averages. H1--H3 denote task groups defined by manipulation-step count, with the corresponding tasks shown on the right. TemporalFlow-VLA achieves the best overall performance, with the clearest gains on longer-horizon, multi-stage tasks. \textbf{Center:} our method augments a standard VLM backbone with a lightweight temporal module alongside the action branch. \textbf{Right:} representative predicted robot-surface temporal-flow fields from the same 12 tasks, grouped by manipulation horizon; arrows visualize motion direction and magnitude on the robot surface.}
    \label{fig:teaser}
\end{minipage}
\vspace{0.4em}
}

\begin{document}
\maketitle
\thispagestyle{empty}
\pagestyle{empty}

\begin{abstract}
Vision-language-action (VLA) models leverage pretrained vision-language representations for robot control, yet simply adding historical frames does not reliably capture recent physical change. This is especially problematic in multi-stage manipulation, where visually similar states may require different actions depending on prior execution. To address this challenge, we present TemporalFlow-VLA, which learns compact execution history through physically grounded temporal supervision. Using recorded robot states, robot geometry, and calibrated cameras, we construct robot-surface temporal flow as a training-only target and supervise two execution-aligned temporal queries that provide structured history to the action expert. The geometric supervision path is not evaluated at deployment. TemporalFlow-VLA achieves $\boldsymbol{97.63\pm0.26\%}$ average success on LIBERO, including $\boldsymbol{96.60\pm0.87\%}$ on LIBERO Long, and $\boldsymbol{85.5\%/84.2\%}$ Clean/Randomized success across 12 RoboTwin tasks. It shows its clearest advantage over prior methods on longer-horizon, multi-stage manipulation. Controlled history interventions show that action prediction depends on both historical content and temporal order. With asynchronous feature caching, temporal conditioning maintains single-frame-level server-side sampling latency without additional historical-encoding overhead. Overall, TemporalFlow-VLA provides a compact, physically grounded interface for exploiting ordered execution history without explicit motion estimation or geometric processing at deployment.
\end{abstract}

\section{INTRODUCTION}

Vision-language-action (VLA) models transfer pretrained vision-language representations to robot control and have shown strong generalization across tasks and embodiments \cite{openvla,pi0,pi05}. Yet many representative VLAs still generate each action chunk primarily from the current RGB observation, language instruction, and robot state. This becomes ambiguous when visually similar observations arise from different execution histories: a failed grasp may resemble the pre-grasp state, and the same end-effector pose can correspond to approaching, carrying, or recovering. Without recent temporal evolution, a policy may misidentify the local task phase, overlook the outcome of the preceding chunk, or repeat already executed behavior, with errors accumulating across replanning steps.

\begin{figure*}[t]
    \centering
    \includegraphics[width=0.88\textwidth]{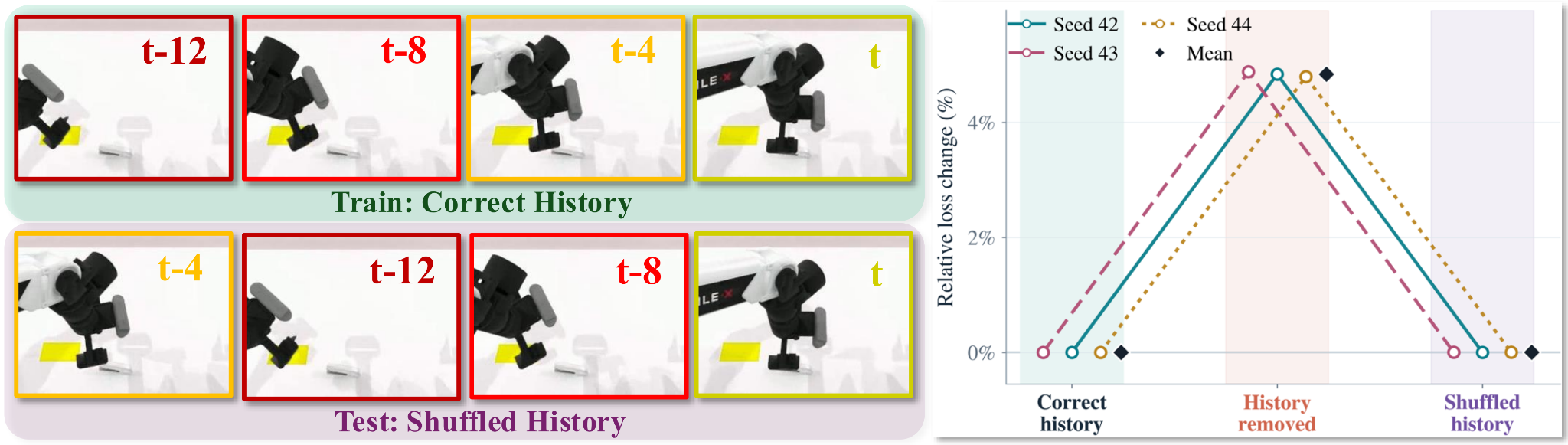}
    \caption{History-use diagnostic. The baseline is trained with chronologically ordered history. At evaluation, shuffling the three historical observations while fixing the current frame leaves the offline action flow-matching loss nearly unchanged, whereas removing history increases it by about 4.6\%.}
    \label{fig:history_diagnostic}
\end{figure*}

A straightforward remedy is to provide historical observations and let the VLM backbone or fusion module infer temporal cues. Yet more frames do not guarantee an action-usable representation of physical change. Figure~\ref{fig:teaser} summarizes our key observation and approach: physically grounded robot-surface temporal flow provides an interpretable training signal for learning compact temporal queries, while the resulting representation delivers stronger benefits as manipulation horizons increase. ViSTR-Bench reports persistent difficulty in motion perception, spatial relations, outcome prediction, and physical dynamics from continuous visual cues \cite{vistr}, while a mechanistic audit of frozen multi-frame VLAs finds that history-unique information is weak and often affects actions mainly when the current observation is unreliable \cite{presentnotremembered}. Our diagnostic shows the same pattern (Fig.~\ref{fig:history_diagnostic}): shuffling three historical frames while fixing the current frame leaves offline action flow-matching loss nearly unchanged, whereas removing history increases it by about 4.6\%. Thus history helps, but this unconstrained baseline is largely insensitive to its correct order. TraceVLA provides an important counterpoint: converting tracked point trajectories into visual prompts explicitly exposes historical motion to the VLA and substantially outperforms a six-frame history baseline \cite{tracevla}. Together, these results suggest that the key challenge is not simply providing history, but representing recent physical evolution in a form structured for control.

To address this challenge, we introduce TemporalFlow-VLA, which uses physically grounded motion as supervision rather than as an additional inference-time prompt. A parallel temporal module learns from robot-surface motion projected into the policy image using recorded joint states, URDF geometry, and camera calibration. Given RGB observations at $t-15$, $t-8$, and $t$, two compact temporal queries, $Q_{15}$ and $Q_{8}$, are supervised at chunk-aligned temporal scales and are the only historical representations exposed to the action expert through joint masked attention. This differs from motion-prompting approaches such as TraceVLA, which estimate image-space trajectories and feed the resulting visual traces to the policy at test time \cite{tracevla}: in our method, dense temporal flow is only a training target, and the geometric supervision path is not evaluated at deployment. The policy retains only sparse historical observations/features, with asynchronous caching to limit latency. TemporalFlow-VLA reaches $97.63\pm0.26\%$ average success on LIBERO and 84.2\% across 12 challenging randomized RoboTwin 2.0 tasks \cite{libero,robotwin2}, with its clearest gains on long- and multi-stage manipulation.

\begin{figure*}[!t]
    \centering
    \includegraphics[width=\textwidth]{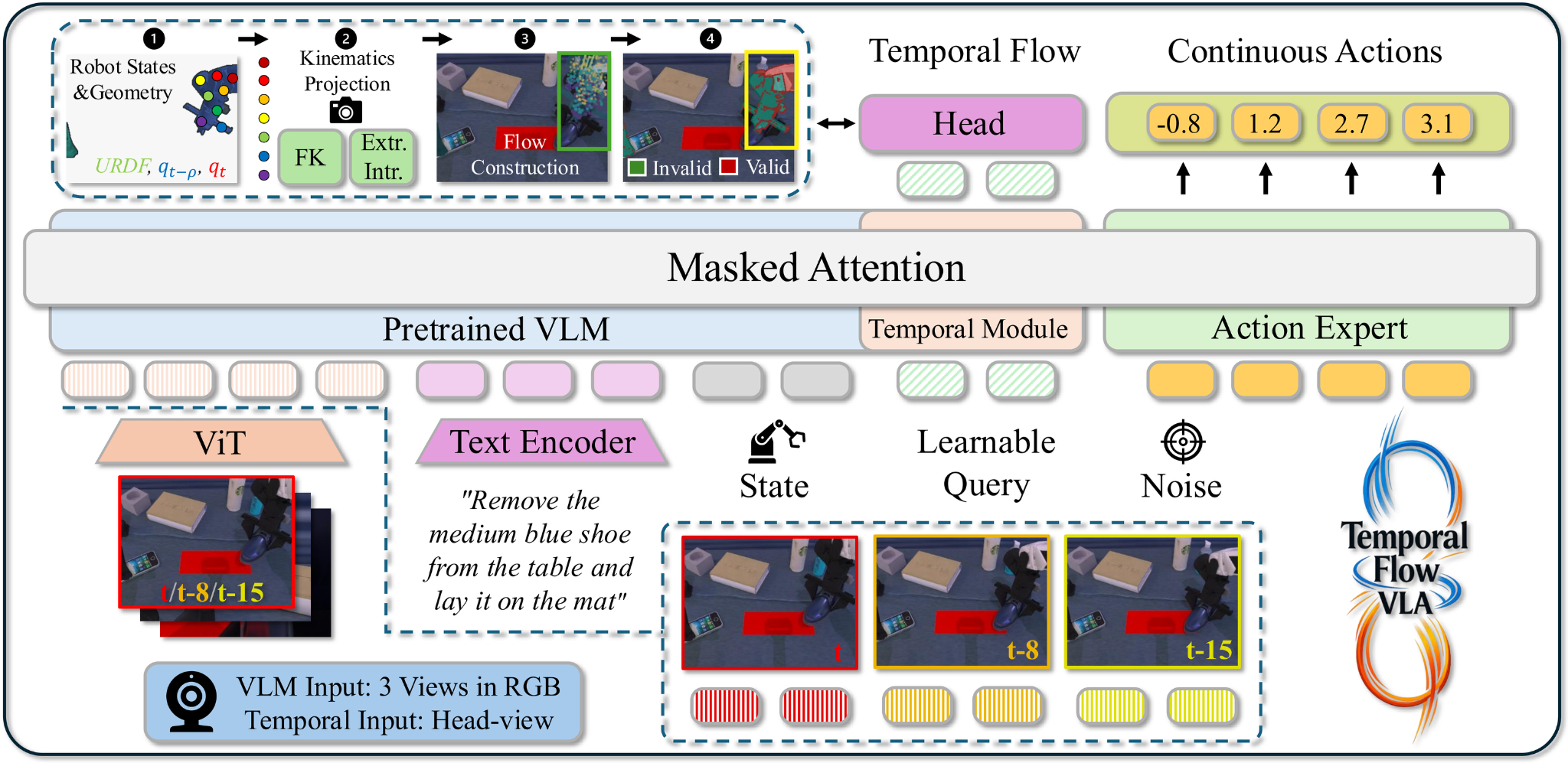}
    \caption{Overview of TemporalFlow-VLA. Robot state, URDF geometry, and calibrated camera parameters generate robot-surface temporal-flow supervision offline. A parallel temporal module compresses historical RGB observations into flow-supervised temporal queries, which are exposed to the action expert through the model's joint masked-attention operation.}
    \label{fig:overview}
\end{figure*}

Our contributions are fourfold. \textbf{First,} we introduce kinematics-grounded robot-surface temporal flow, which projects recorded robot motion into the policy RGB plane to provide explicit, interpretable supervision without object annotations or additional deployment-time sensors. \textbf{Second,} we propose two chunk-aligned temporal queries whose physical content is defined by interval-specific flow reconstruction. Historical image patches cannot directly reach the action expert; $Q_{15}$ and $Q_{8}$ form the compact, supervised interface through joint masked attention (Fig.~\ref{fig:overview}). \textbf{Third,} we introduce an asynchronous historical-feature cache that overlaps historical visual encoding with action execution, substantially reducing the inference overhead of temporal conditioning. \textbf{Fourth,} extensive simulation and real-robot evaluations validate the effectiveness and practical feasibility of TemporalFlow-VLA.

\section{RELATED WORK}

\subsection{Vision-Language-Action Models}
Vision-language-action models adapt pretrained vision-language representations to end-to-end robot control \cite{openvla,pi0,pi05}. Recent flow-based VLAs such as $\pi_0$ and $\pi_{0.5}$ generate continuous action chunks with strong generalization \cite{pi0,pi05}, but low-level decisions are still largely conditioned on the present observation. This is ambiguous in multi-stage manipulation, where similar-looking states may follow different motion histories. We preserve the base current-observation pathway while adding a compact representation of recent image-aligned robot motion.

\subsection{External Memory and Explicit Motion}
MemER retrieves historical keyframes to guide a low-level VLA \cite{memer}, MEM combines video-based short- and language-based long-term memory \cite{mem}, and TempoFit reuses layer-wise prefix K/V states with recency-biased retrieval as a training-free temporal retrofit \cite{tempofit}. These methods organize or retrieve temporal context; we instead learn the local execution history immediately preceding replanning and assign it an explicit physical target.

Motion-centric methods make temporal change more explicit. TraceVLA overlays tracked point trajectories as visual prompts \cite{tracevla}; MotionVLA converts past video into scene-wide trajectory-field tokens \cite{motionvla}; and HiF-VLA uses codec motion vectors as hindsight priors together with foresight motion reasoning \cite{hifvla}. FlowVLA instead inserts optical flow as a visual chain-of-thought for future-frame world-model pretraining, $v_t\!\rightarrow\! f_t\!\rightarrow\! v_{t+1}$ \cite{flowvla}. TemporalFlow-VLA uses deterministic robot-surface flow from robot state, URDF geometry, and camera calibration only to supervise compact past-to-current historical queries during training; no flow, trajectory, or motion-vector estimator remains at deployment.

\subsection{Latent Temporal Representations in VLAs}
A second line of work learns compact temporal representations from VLM features. CronusVLA aggregates multi-frame motion features through feature chunking \cite{cronusvla}; HAMLET uses time-contrastive moment tokens and a lightweight memory module \cite{hamlet}; MemoryVLA forms perceptual and cognitive working-memory tokens from the current observation and uses them to retrieve decision-relevant entries from a Perceptual-Cognitive Memory Bank \cite{memoryvla}; and ReMem-VLA propagates frame-level and chunk-level recurrent memory queries \cite{rememvla}. Together, these studies show that latent memories can improve temporally dependent control.

However, their temporal content is largely shaped by native VLM features, action supervision, reconstruction, or recurrence, without prescribing which physical change each latent should encode. This matters because generic multimodal representations do not necessarily provide action-usable representations of continuous dynamics \cite{vistr,presentnotremembered}. TemporalFlow-VLA asks a complementary question: can compact latent history be assigned an explicit, control-aligned physical target? $Q_{15}$ and $Q_{8}$ are supervised to recover robot-surface flow over two action-chunk-aligned intervals, giving each token a defined temporal scale and observable motion semantics while keeping the action interface compact.

\section{METHOD}

\subsection{Overview}
TemporalFlow-VLA introduces a parallel temporal pathway into a pretrained vision-language-action policy while preserving the original action-generation path. The base policy continues to predict an action chunk from the current RGB observation, language instruction, robot state, and diffusion timestep, whereas the temporal pathway additionally receives head-camera observations from $t-15$, $t-8$, and $t$. For a 16-step action chunk, $t-15$ and $t-8$ approximately correspond to the beginning and midpoint of the previous chunk, providing historical context over both the full chunk and its more recent half.

The pathway is built around two supervised temporal queries, $Q_8$ and $Q_{15}$. $Q_8$ represents recent motion from $t-8$ to $t$, while $Q_{15}$ captures the complete evolution from $t-15$ to $t$ and can further build on the short-range summary encoded by $Q_8$. Action tokens obtain historical information only through these two queries and cannot directly access historical image patches. During training, robot kinematics provides robot-surface temporal-flow supervision for the queries. At deployment, the kinematic renderer is absent and flow-reconstruction heads are not evaluated; only temporal-query and action-generation computations remain, with history reused through an asynchronous cache.

\subsection{Kinematics-Grounded Robot-Surface Temporal Flow}
Our goal is to supervise how robot motion appears in RGB. For each interval $\rho\in\{8,15\}$, let $s=t-\rho$. Let $\mathbf{q}_{\tau}$ be the joint configuration, $\mathbf{T}_{B}^{l}(\mathbf{q}_{\tau})$ the homogeneous link-to-base forward-kinematic transform, and $\mathbf{T}_{C\leftarrow B}$ the calibrated base-to-camera transform; overbars denote homogeneous coordinates. A robot-only renderer gives each visible robot source pixel $p$ its 3-D base-frame surface intersection $\mathbf{X}_{s}^{B}(p)$ and owning link $l(p)$. These depth-visible pixels, rather than fixed mesh samples, define the supervision and naturally weight surfaces by projected area. With intrinsics $K$, $\Pi_K$ denotes perspective projection from camera coordinates to renderer pixels. We recover the point in its link frame, transport it with target-time forward kinematics, and project it into the target image:
\begin{align}
\bar{\mathbf{x}}^{l(p)}(p) &= \left[\mathbf{T}_{B}^{l(p)}(\mathbf{q}_{s})\right]^{-1}\bar{\mathbf{X}}_{s}^{B}(p), \\
\bar{\mathbf{X}}_{t}^{B}(p) &= \mathbf{T}_{B}^{l(p)}(\mathbf{q}_{t})\,\bar{\mathbf{x}}^{l(p)}(p), \\
\tilde{\mathbf{u}}_{t}(p) &= \Pi_K\!\left(\mathbf{T}_{C\leftarrow B}\bar{\mathbf{X}}_{t}^{B}(p)\right).
\label{eq:projection}
\end{align}
Nearest-pixel lookup in a target robot-only position/link-ID render retains a correspondence only when the projection is in bounds, belongs to the same link $l(p)$, and has a 3-D residual no larger than $5\,\mathrm{mm}$.

Let $\mathbf{u}_{s}(p)$ and $\mathbf{u}_{t}(p)$ denote the source and valid transported target locations after both are expressed on the $224\!\times\!224$ policy image plane. We normalize displacement by the image size,
\begin{equation}
\mathbf{f}_{\rho}(p)=\frac{\mathbf{u}_{t}(p)-\mathbf{u}_{s}(p)}{224}.
\label{eq:normalizedflow}
\end{equation}
Masked arithmetic mean over non-overlapping $14\!\times\!14$ regions converts valid robot flows into a $16\!\times\!16\!\times\!2$ target. For coverage, $\mathcal{S}_k$ contains all source-image pixels in patch $k$ (including background), while $\mathcal{V}_k\subseteq\mathcal{S}_k$ contains only valid self-visible robot pixels. For patches with valid support,
\begin{align}
\mathbf{F}_{\rho}(k) &= \frac{1}{|\mathcal{V}_k|}\sum_{p\in\mathcal{V}_k}\mathbf{f}_{\rho}(p), \\
c_{\rho}(k) &= \frac{|\mathcal{V}_k|}{|\mathcal{S}_k|},\qquad \mathbf{M}_{\rho}(k)=\mathbb{1}[c_{\rho}(k)\ge 0.1].
\label{eq:patchflow}
\end{align}
Thus $c_{\rho}(k)$ measures valid robot support over the full patch area, not robot-conditional coverage; it gates the loss but is not a loss weight. The robot-surface restriction applies only to this auxiliary target: the temporal pathway still receives full RGB, so object and scene changes remain available to the action objective. Labels are generated offline from robot states, geometry, and calibration, without manual flow annotation or deployment-time geometry.

\subsection{Hierarchical Temporal Queries and Joint Attention}
The learnable temporal tokens $\mathbf{q}_{8}$ and $\mathbf{q}_{15}$, corresponding to $Q_8$ and $Q_{15}$, are appended to the standard $\pi_{0.5}$ prefix $\mathbf{P}_{\mathrm{std}}(\mathbf{V}_{0},\mathbf{L},\mathbf{s})$, where $\mathbf{V}_{0}$, $\mathbf{L}$, and $\mathbf{s}$ denote current-image tokens, language tokens, and the current robot-state input. Learned frame-identity embeddings are added only to historical patches $\mathbf{V}_{15}$ and $\mathbf{V}_{8}$:
\begin{equation}
\mathbf{P}^{0}=\left[\mathbf{P}_{\mathrm{std}}(\mathbf{V}_{0},\mathbf{L},\mathbf{s});\mathbf{V}_{15};\mathbf{V}_{8};\mathbf{q}_{8};\mathbf{q}_{15}\right].
\label{eq:prefix}
\end{equation}

Q8 and Q15 are not treated as independent queries. Instead, a directed query-specific mask organizes them into a hierarchy. $Q_8$ can read only the language tokens, the $t-8$ observation, the current observation, and itself, and therefore specializes in recent motion. $Q_{15}$ can additionally read the $t-15$ observation and $Q_8$, allowing it to integrate earlier history on top of the short-range summary:
\begin{align}
\mathcal{A}_{8} &= \{\mathbf{L},\mathbf{V}_{8},\mathbf{V}_{0},\mathbf{q}_{8}\}, \\
\mathcal{A}_{15} &= \{\mathbf{L},\mathbf{V}_{15},\mathbf{V}_{8},\mathbf{V}_{0},\mathbf{q}_{8},\mathbf{q}_{15}\}, \\
Q_8 &\longrightarrow Q_{15}.
\label{eq:querymask}
\end{align}

\begin{figure}[t]
    \centering
    \includegraphics[width=0.82\columnwidth]{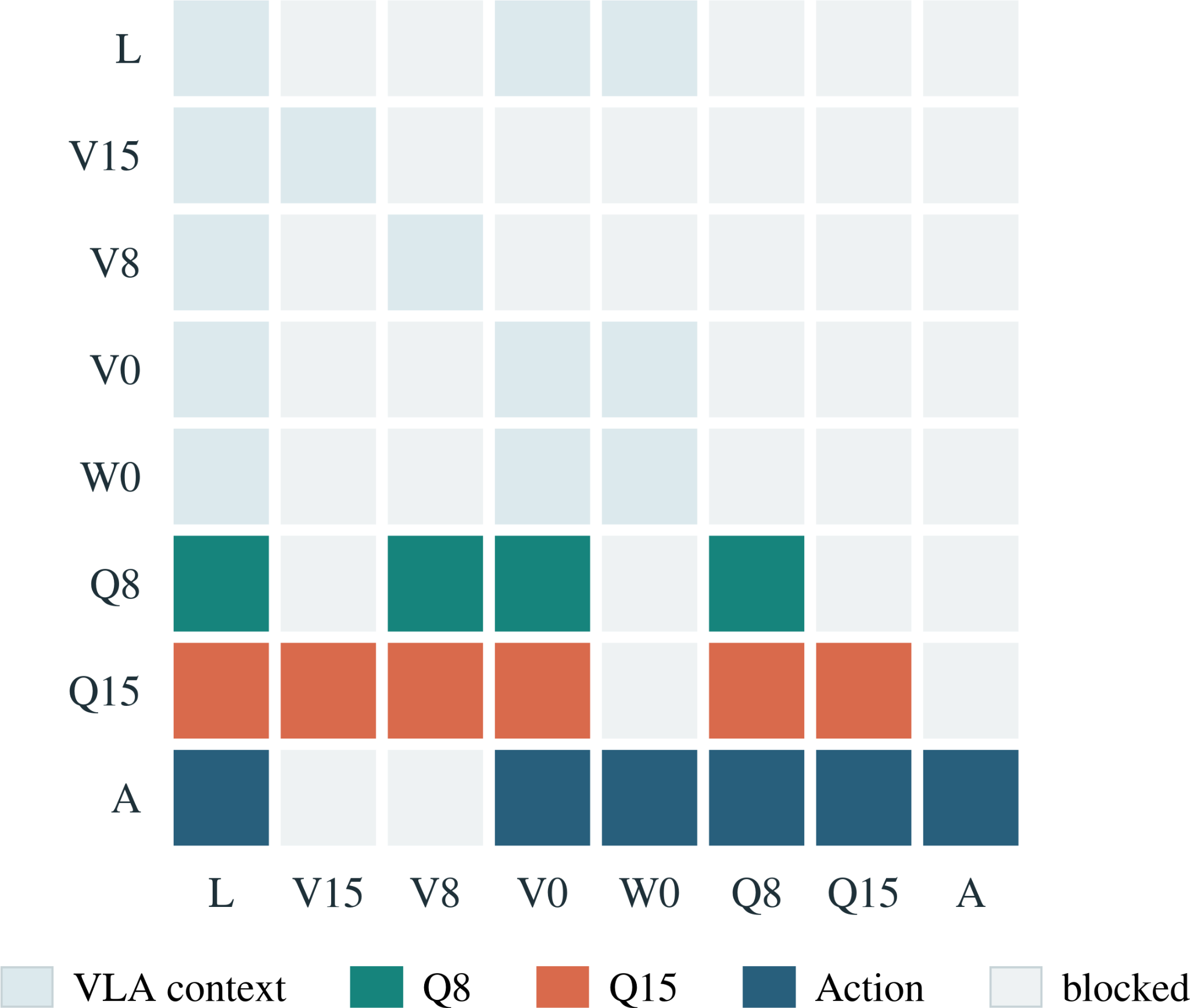}
    \caption{Directed masked-attention pattern. $Q_8$ summarizes short-range history, $Q_{15}$ may additionally read $Q_8$ and the longer-range frame, and action tokens access historical visual information only through the two temporal queries.}
    \label{fig:mask}
\end{figure}

The one-way $Q_8\!\rightarrow\!Q_{15}$ path lets $Q_{15}$ integrate long-range context over $Q_8$'s recent-motion summary while keeping the two temporal scales distinct (Fig.~\ref{fig:mask}).

Let $\mathbf{z}_{\rho}\in\mathbb{R}^{2048}$ denote the final state of $Q_{\rho}$ and $\mathbf{V}_{\rho}\in\mathbb{R}^{16\times16\times2048}$ the final VLM spatial patch states of its source frame $I_{t-\rho}$. Each interval uses a separate query-conditioned decoder $D_{\rho}(\mathbf{V}_{\rho},\mathbf{z}_{\rho})$: the query FiLM-modulates the source spatial states and a pointwise two-layer MLP predicts flow,
\begin{align}
\mathbf{H}_{\rho} &= \operatorname{LN}(\mathbf{V}_{\rho})\odot\left(1+\boldsymbol{\gamma}_{\rho}(\mathbf{z}_{\rho})\right)+\boldsymbol{\beta}_{\rho}(\mathbf{z}_{\rho}), \\
\hat{\mathbf{F}}_{\rho} &= \mathbf{W}_{\rho}^{o}\,\operatorname{GELU}\!\left(\mathbf{W}_{\rho}^{h}\mathbf{H}_{\rho}+\mathbf{b}_{\rho}^{h}\right)+\mathbf{b}_{\rho}^{o}, \quad \rho\in\{8,15\}.
\label{eq:filmdecoder}
\end{align}
Here, $\boldsymbol{\gamma}_{\rho}$ and $\boldsymbol{\beta}_{\rho}$ are learned channel-wise FiLM projections of $\mathbf{z}_{\rho}$, and $\mathbf{W}_{\rho}^{h},\mathbf{W}_{\rho}^{o},\mathbf{b}_{\rho}^{h},\mathbf{b}_{\rho}^{o}$ parameterize the interval-specific pointwise MLP. The two decoders share no parameters and use no convolution or upsampling. With $\ell_{\mathrm{Huber}}$ defined as the mean over the two flow coordinates, loss is evaluated only where $\mathbf{M}_{\rho}=1$:
\begin{align}
\mathcal{L}_{\mathrm{flow}}^{\rho} &= \frac{\sum_k \mathbf{M}_{\rho}(k)\,\ell_{\mathrm{Huber}}\!\left(\hat{\mathbf{F}}_{\rho}(k),\mathbf{F}_{\rho}(k)\right)}{\sum_k \mathbf{M}_{\rho}(k)+\epsilon}, \\
\mathcal{L}_{\mathrm{temp}} &= \frac{1}{2}\left(\mathcal{L}_{\mathrm{flow}}^{8}+\mathcal{L}_{\mathrm{flow}}^{15}\right),
\label{eq:temploss}
\end{align}
where $\epsilon>0$ is a small constant for numerical stability.

Temporal information enters the action expert through the model's existing joint masked self-attention. At layer $l$, expert-specific projections form queries, keys, and values from prefix $\mathbf{P}^{l}$ and action suffix $\mathbf{A}^{l}$, which are then concatenated:
\begin{align}
\mathbf{Q}^{l} &= [\mathbf{P}^{l}\mathbf{W}_{Q,p}^{l};\mathbf{A}^{l}\mathbf{W}_{Q,a}^{l}], \\
\mathbf{K}^{l} &= [\mathbf{P}^{l}\mathbf{W}_{K,p}^{l};\mathbf{A}^{l}\mathbf{W}_{K,a}^{l}], \\
\mathbf{V}^{l} &= [\mathbf{P}^{l}\mathbf{W}_{V,p}^{l};\mathbf{A}^{l}\mathbf{W}_{V,a}^{l}], \\
\mathbf{O}^{l} &= \operatorname{Softmax}\!\left(\frac{\mathbf{Q}^{l}(\mathbf{K}^{l})^{\top}}{\sqrt{d}}+\mathbf{B}(\mathbf{M})\right)\mathbf{V}^{l}.
\label{eq:jointattn}
\end{align}
Here, $d$ is the per-head query/key dimension, $\mathbf{M}$ is the structured attention mask, and $\mathbf{B}(\mathbf{M})$ converts disallowed connections into negative-infinity attention biases. For action tokens, the mask preserves the original VLA context and additionally exposes $Q_8$ and $Q_{15}$, while blocking direct access to $\mathbf{V}_{8}$ and $\mathbf{V}_{15}$. Historical visual information must therefore be compressed into flow-supervised query representations before it can influence action generation. This design requires neither a separate cross-attention module nor a history-specific residual gate and preserves the action expert's original AdaRMS residual modulation.

The final objective jointly optimizes the original action flow-matching loss and the temporal-flow reconstruction loss:
\begin{equation}
\mathcal{L}=\mathcal{L}_{\mathrm{action}}+\lambda_{\mathrm{temp}}\mathcal{L}_{\mathrm{temp}}.
\label{eq:objective}
\end{equation}
where $\lambda_{\mathrm{temp}}$ controls the auxiliary temporal-flow objective; we set $\lambda_{\mathrm{temp}}=1.0$ in all main experiments. Maintaining this supervision throughout training keeps $Q_8$ and $Q_{15}$ tied to their intended temporal scales and motion semantics rather than unconstrained historical latents.

\begin{table*}[!t]
\centering
\caption{RoboTwin 2.0 success rate (\%), grouped by execution horizon following LingBot-VA \cite{lingbotva}. The $\pi_0$ and $\pi_{0.5}$ task-level entries are from LingBot-VA (Easy/Hard). X-VLA$^{\dagger}$ and GigaWorld entries are from Table~8 of GigaWorld-Policy \cite{gigaworld} (Clean/Rand.); X-VLA$^{\dagger}$ therefore denotes the RoboTwin re-evaluation reported there rather than the original X-VLA results \cite{xvla}. The corresponding fixed/clean and randomized conditions are displayed as Clean/Rand. Best and second-best results are bold and underlined separately within each condition.}
\label{tab:robotwin_main}
\scriptsize
\setlength{\tabcolsep}{1.7pt}
\renewcommand{\arraystretch}{1.03}
\begin{tabular*}{\textwidth}{@{\extracolsep{\fill}}clcccccccccc@{}}
\toprule
\multirow{2}{*}{H.} & \multirow{2}{*}{Simulation Task}
& \multicolumn{2}{c}{\textbf{Ours}}
& \multicolumn{2}{c}{$\pi_0$}
& \multicolumn{2}{c}{$\pi_{0.5}$}
& \multicolumn{2}{c}{X-VLA$^{\dagger}$~\cite{xvla}}
& \multicolumn{2}{c}{GigaWorld} \\
\cmidrule(lr){3-4}\cmidrule(lr){5-6}\cmidrule(lr){7-8}\cmidrule(lr){9-10}\cmidrule(lr){11-12}
& & Clean & Rand. & Clean & Rand. & Clean & Rand. & Clean & Rand. & Clean & Rand. \\
\midrule
\multirow{5}{*}{1}
& Place Shoe           & \underline{97} & \underline{95} & 76 & 76 & 92 & 93 & 96 & \underline{95} & \textbf{98} & \textbf{96} \\
& Place Phone Stand    & 79 & \underline{82} & 49 & 53 & 81 & 81 & \textbf{88} & \textbf{87} & \underline{82} & 72 \\
& Lift Pot             & \underline{98} & \underline{98} & 80 & 72 & 96 & 85 & \textbf{99} & \textbf{100} & \underline{98} & \underline{98} \\
& Move Stapler Pad     & 62 & 62 & 41 & 24 & 56 & 42 & \underline{78} & \underline{73} & \textbf{92} & \textbf{82} \\
& \textit{Average}     & 84.0 & 84.3 & 61.5 & 56.3 & 81.3 & 75.3 & \underline{90.3} & \textbf{88.8} & \textbf{92.5} & \underline{87.0} \\
\midrule
\multirow{5}{*}{2}
& Stack Blocks Two     & \textbf{100} & \textbf{100} & 93 & 79 & \underline{97} & \textbf{100} & 92 & 87 & \textbf{100} & \underline{94} \\
& Place Burger Fries   & \underline{95} & 91 & 81 & 76 & 94 & 87 & 94 & \underline{94} & \textbf{98} & \textbf{96} \\
& Hanging Mug          & \textbf{39} & \textbf{33} & 14 & 11 & 18 & 17 & \underline{23} & \underline{27} & 16 & 12 \\
& Handover Mic         & \textbf{100} & \textbf{99} & 97 & \underline{97} & \underline{98} & \underline{97} & 0 & 0 & 72 & 72 \\
& \textit{Average}     & \textbf{83.5} & \textbf{80.8} & 71.3 & 65.8 & \underline{76.8} & \underline{75.3} & 52.3 & 52.0 & 71.5 & 68.5 \\
\midrule
\multirow{5}{*}{3}
& Stack Blocks Three   & \textbf{97} & \textbf{95} & 72 & 52 & \underline{91} & 76 & 6 & 10 & 70 & \underline{78} \\
& Put Bottles Dustbin  & \textbf{91} & \textbf{86} & 65 & 56 & \underline{84} & \underline{79} & 74 & 77 & 72 & 70 \\
& Blocks Ranking Size  & \textbf{71} & \underline{71} & 14 & 5 & 49 & 26 & \underline{67} & \textbf{74} & 44 & 48 \\
& Blocks Ranking RGB   & \textbf{97} & \textbf{98} & 80 & 63 & \underline{92} & 85 & 83 & 83 & \underline{92} & \underline{96} \\
& \textit{Average}     & \textbf{89.0} & \textbf{87.5} & 57.8 & 44.0 & \underline{79.0} & 66.5 & 57.5 & 61.0 & 69.5 & \underline{73.0} \\
\midrule
\multicolumn{2}{l}{\textbf{Overall Average}} & \textbf{85.5} & \textbf{84.2} & 63.5 & 55.3 & \underline{79.0} & 72.3 & 66.7 & 67.3 & 77.8 & \underline{76.2} \\
\bottomrule
\end{tabular*}
\end{table*}

\begin{table}[!t]
\centering
\caption{LIBERO success rate (\%). Ours reports mean $\pm$ SD over three seeds; published methods are shown as reported in their sources, with GR00T N1.7 percentages recomputed from the official success counts \cite{openpi_libero,groot_libero}.}
\label{tab:libero_main}
\scriptsize
\setlength{\tabcolsep}{2.3pt}
\renewcommand{\arraystretch}{0.98}
\resizebox{\columnwidth}{!}{%
\begin{tabular}{lccccc}
\toprule
Method & Spatial & Object & Goal & Long & Avg. \\
\midrule
$\pi_{0.5}$ \cite{pi05} & \underline{98.8} & 98.2 & \underline{98.0} & 92.4 & 96.85 \\
GR00T N1.7 \cite{groot_libero} & 97.5 & 98.5 & 97.5 & \underline{94.5} & 97.0 \\
CronusVLA \cite{cronusvla} & 97.3 & \underline{99.6} & 96.9 & 94.0 & 97.0 \\
HAMLET \cite{hamlet} & \textbf{99.0} & \textbf{100.0} & \textbf{99.2} & 92.2 & \textbf{97.7} \\
MotionVLA \cite{motionvla} & 96.2 & 98.0 & 96.2 & 91.2 & 95.4 \\
\midrule
\textbf{Ours} & $97.60{\pm}0.20$ & $99.40{\pm}0.20$ & $96.93{\pm}0.61$ & $\mathbf{96.60{\pm}0.87}$ & \underline{$97.63{\pm}0.26$} \\
\bottomrule
\end{tabular}}
\end{table}

\subsection{Asynchronous Historical-Feature Caching}
\begin{figure}[H]
    \centering
    \includegraphics[width=0.72\columnwidth]{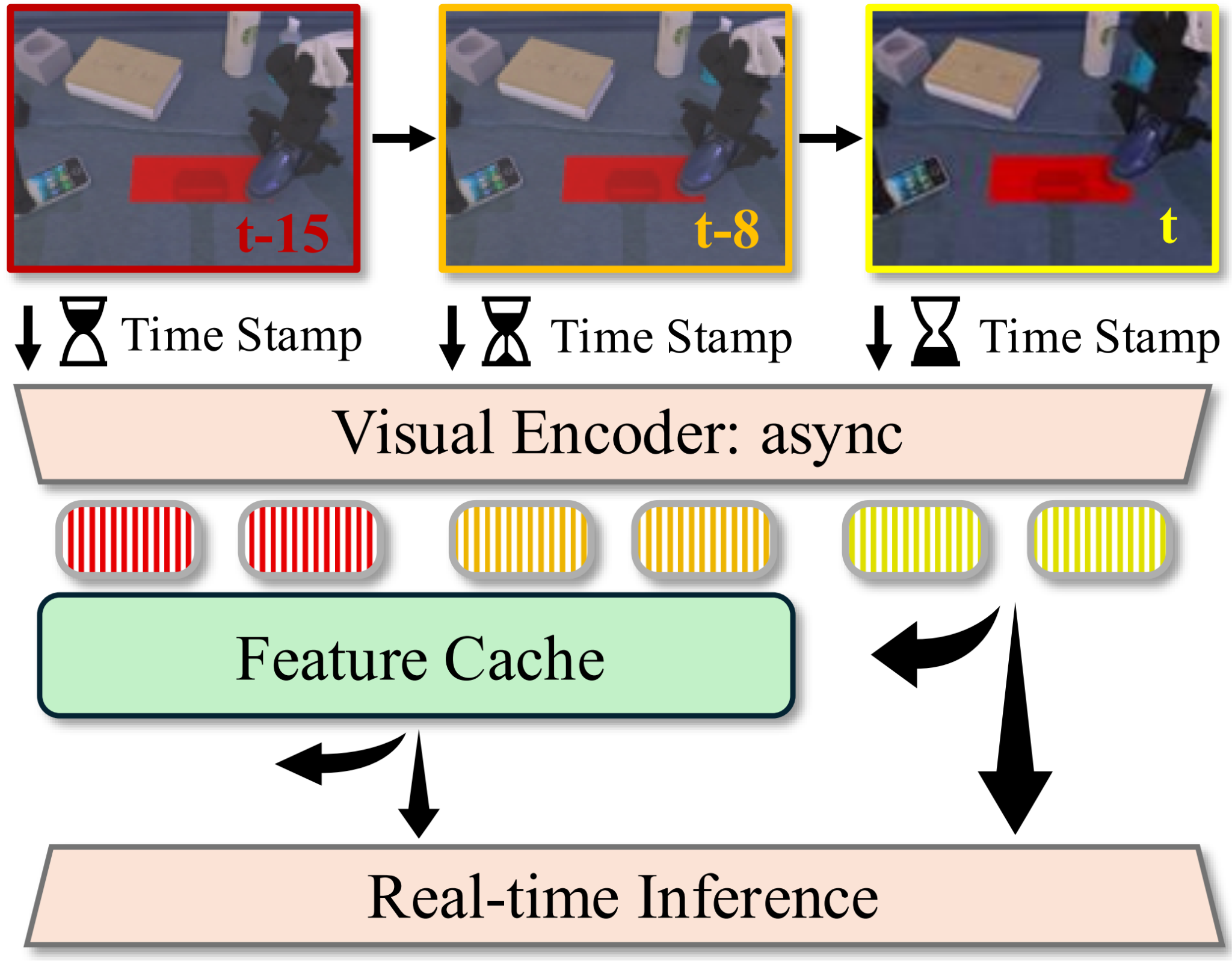}
    \caption{Asynchronous historical-feature caching. Historical observations are encoded during execution and stored in a timestamped feature cache, so replanning retrieves cached history while only the current observation remains on the synchronous inference path.}
    \label{fig:cache}
\end{figure}

Synchronously encoding the $t-15$, $t-8$, and $t$ observations at every replanning step would place historical image encoding on the inference critical path and introduce two additional visual forward passes. To avoid this cost (Fig.~\ref{fig:cache}), incoming head-camera observations are written into a timestamped ring buffer while the robot executes the current action chunk, and a background process asynchronously extracts and stores their visual features:
\begin{align}
\mathbf{V}_{\tau} &= \operatorname{VisualEncoder}(I_{\tau}), \\
\mathcal{C}[\tau] &= \mathbf{V}_{\tau}.
\label{eq:cachewrite}
\end{align}
Because the action-chunk length and the $Q_8/Q_{15}$ offsets are fixed, the $t-15$ and $t-8$ observations required at the next replanning step can be encoded during execution of the current chunk. At replanning time $t$, the two historical features are retrieved directly from the cache, whereas the current observation is encoded synchronously and shared with the original VLA context:
\begin{align}
\mathbf{V}_{15} &= \mathcal{C}[t-15], \\
\mathbf{V}_{8} &= \mathcal{C}[t-8], \\
\mathbf{P}_{\mathrm{temp}} &= [\mathbf{V}_{15}^{\mathrm{cache}};\mathbf{V}_{8}^{\mathrm{cache}};\mathbf{V}_{0}^{\mathrm{current}}].
\label{eq:cacheread}
\end{align}
Without caching, synchronous latency includes three visual encodings. With asynchronous caching, the two historical encodings overlap with execution of the previous action chunk, leaving only current-frame encoding, the joint transformer, and action generation on the critical path:
\begin{align}
T_{\mathrm{naive}} &= 3T_{\mathrm{vision}}+T_{\mathrm{joint}}+T_{\mathrm{action}}, \\
T_{\mathrm{cache}} &\approx T_{\mathrm{vision}}+T_{\mathrm{joint}}+T_{\mathrm{action}}.
\label{eq:latency}
\end{align}
Here, $T_{\mathrm{vision}}$, $T_{\mathrm{joint}}$, and $T_{\mathrm{action}}$ denote visual-encoding, joint-transformer (including temporal queries), and action-generation latency.

The cache removes only synchronous historical-image encoding; temporal queries remain in the joint transformer. At episode start, unavailable history is filled with the earliest observation and marked invalid; the cache is bypassed until both required historical frame tokens are available. Stale features are evicted outside the required window.

\section{EXPERIMENTS}

\subsection{Experimental Setup}
Our experiments consist of simulation benchmark evaluation, controlled ablations, inference-efficiency evaluation, and real-robot experiments. All policies are trained on 8 NVIDIA H100 GPUs with a per-GPU batch size of 32. On LIBERO \cite{libero}, we jointly train LIBERO Spatial, Object, Goal, and Long for 30k steps and report mean $\pm$ SD over seeds 0, 2, and 5, with 500 rollouts per suite and seed. On RoboTwin 2.0 \cite{robotwin2}, we jointly train 12 tasks for 60k steps from 50 clean and 500 randomized demonstrations per task, then evaluate 100 rollouts per task in each setting and group tasks by execution horizon following LingBot-VA \cite{lingbotva}. Temporal-flow labels are precomputed offline; under the same hardware and batch size, adding the temporal module and auxiliary flow decoders increases wall-clock training time by approximately 20\% relative to the baseline. At deployment, the geometric label pipeline is absent and the auxiliary flow decoders are not evaluated.

\subsection{Simulation Benchmark Evaluation}

RoboTwin is our primary long-horizon evaluation. As shown in Table~\ref{tab:robotwin_main}, TemporalFlow-VLA reaches 85.5\% and 84.2\% average SR under clean and randomized evaluation, respectively. Under the randomized setting, our method exceeds the best reported result among the published baselines included in our comparison by 8.0 percentage points. More importantly, the advantage grows with task horizon: our method obtains 80.8\% at $H{=}2$ and 87.5\% at $H{=}3$, exceeding the respective runner-up averages by 5.5 and 14.5 points. In contrast, the $H{=}1$ average is not the best. This horizon-dependent pattern is consistent with our motivation: explicit recent execution history is most useful when success depends on maintaining progress across multiple sequential stages.

Table~\ref{tab:libero_main} shows complementary evidence on the standard LIBERO suites. TemporalFlow-VLA averages $97.63\pm0.26\%$ over three seeds and remains near the saturated performance of the strongest published systems. Its clearest advantage appears on LIBERO Long, where it reaches $96.60\pm0.87\%$, 2.1 percentage points above the strongest listed prior mean result. The improvement is therefore concentrated in the regime most aligned with our objective---multi-stage manipulation that benefits from knowing how the current state was reached---rather than in already saturated short-horizon suites.

\subsection{Ablation Studies}
\label{sec:ablation}
\begin{figure}[!t]
    \centering
    \includegraphics[width=\columnwidth]{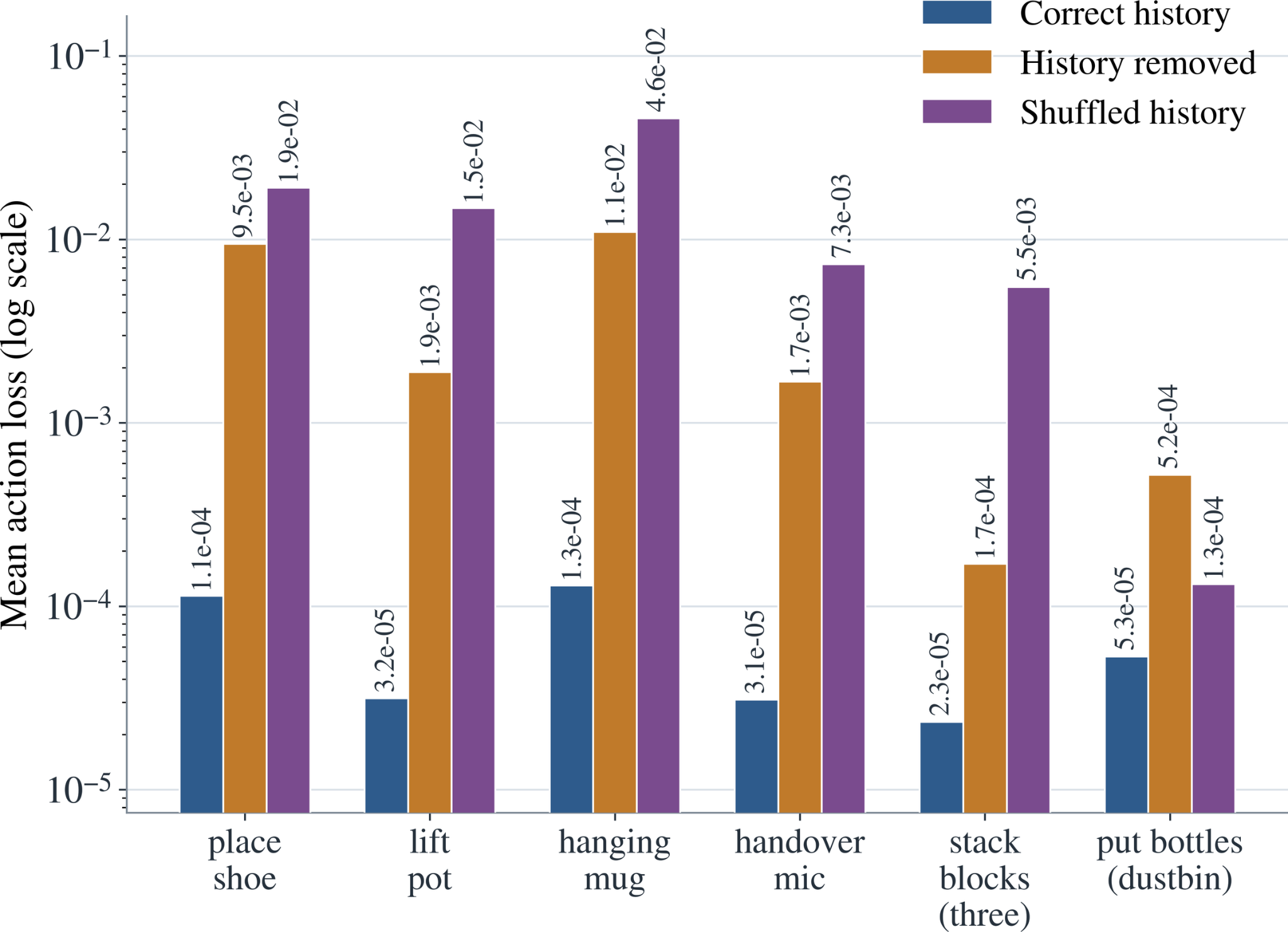}
    \caption{Offline history perturbations on six RoboTwin tasks. Both removing and shuffling history increase action flow-matching loss relative to correct history; shuffling is most harmful on five tasks, while \emph{put bottles (dustbin)} is more sensitive to history removal. The vertical axis is logarithmic.}
    \label{fig:history_order_ours}
\end{figure}

We test history content and order on six RoboTwin tasks using matched windows and diffusion noise. Figure~\ref{fig:history_order_ours} compares correct, removed, and shuffled history (swapping $t-15$ and $t-8$). Both perturbations increase action flow-matching loss on all tasks; shuffling is worst on five, while \emph{put bottles (dustbin)} is more sensitive to removal. This is an offline action-loss diagnostic showing sensitivity to temporal assignment, not a proxy for online success.

Table~\ref{tab:online_ablation} disentangles raw history, the query bottleneck, flow supervision, and temporal scale. Two queries without flow outperform Multi-frame overall (83.2\%/83.6\% vs. 79.8\%/81.5\% on Clean/Randomized), while adding flow reaches 85.5\%/84.2\% and adds 3.7/1.5 points at $H{=}2$. $Q_8$ alone outperforms $Q_{15}$ overall (84.8\%/84.0\% vs. 83.9\%/83.0\%), but both supervised scales are best overall, indicating complementary longer-range context from $Q_{15}$. Because 2Q w/o Flow keeps the same two query slots and history window as Ours, the remaining gap isolates flow supervision from temporal capacity and context length.

\begin{table}[!t]
\centering
\caption{Online RoboTwin ablation by horizon (SR, \%). \emph{Multi-frame}: raw history; \emph{2Q w/o Flow}: two queries without flow supervision. Single-query variants use one supervised query. Bold/underline: best/second.}
\label{tab:online_ablation}
\vspace{-0.3em}
\scriptsize
\setlength{\tabcolsep}{2.8pt}
\renewcommand{\arraystretch}{0.76}
\begin{tabular}{lcccc}
\toprule
Method & $H{=}1$ & $H{=}2$ & $H{=}3$ & Avg. \\
\midrule
\multicolumn{5}{l}{\textit{Clean}} \\
$\pi_{0.5}$ & 81.3 & 76.8 & 79.0 & 79.0 \\
Multi-frame & 75.5 & 79.5 & 84.5 & 79.8 \\
2Q w/o Flow & 83.0 & 79.8 & 86.8 & 83.2 \\
$Q_8$ only & \textbf{86.5} & 79.8 & \underline{88.3} & \underline{84.8} \\
$Q_{15}$ only & \underline{84.0} & \underline{80.5} & 87.3 & 83.9 \\
\textbf{Ours} & \underline{84.0} & \textbf{83.5} & \textbf{89.0} & \textbf{85.5} \\
\midrule
\multicolumn{5}{l}{\textit{Randomized}} \\
$\pi_{0.5}$ & 75.3 & 75.3 & 66.5 & 72.3 \\
Multi-frame & 82.8 & 75.8 & 86.0 & 81.5 \\
2Q w/o Flow & \textbf{85.5} & 79.3 & 86.0 & 83.6 \\
$Q_8$ only & 83.8 & \underline{79.5} & \textbf{88.8} & \underline{84.0} \\
$Q_{15}$ only & 82.5 & 79.0 & \underline{87.5} & 83.0 \\
\textbf{Ours} & \underline{84.3} & \textbf{80.8} & \underline{87.5} & \textbf{84.2} \\
\bottomrule
\end{tabular}
\vspace{-0.35em}
\end{table}

\FloatBarrier

\subsection{Inference Efficiency}

History-conditioned policies repeatedly re-encode overlapping past observations across replans. We therefore cache the frozen visual tokens of the two historical head frames and precompute them asynchronously during execution of the preceding 16-step action chunk. On an RTX 4090, we compare matched successful LIBERO Long rollouts with and without caching, measuring server-side policy sampling time over the first 15 replans after warm-up and excluding RPC communication, simulator stepping, and video writing.

\begin{figure}[!t]
    \centering
    \includegraphics[width=0.88\columnwidth]{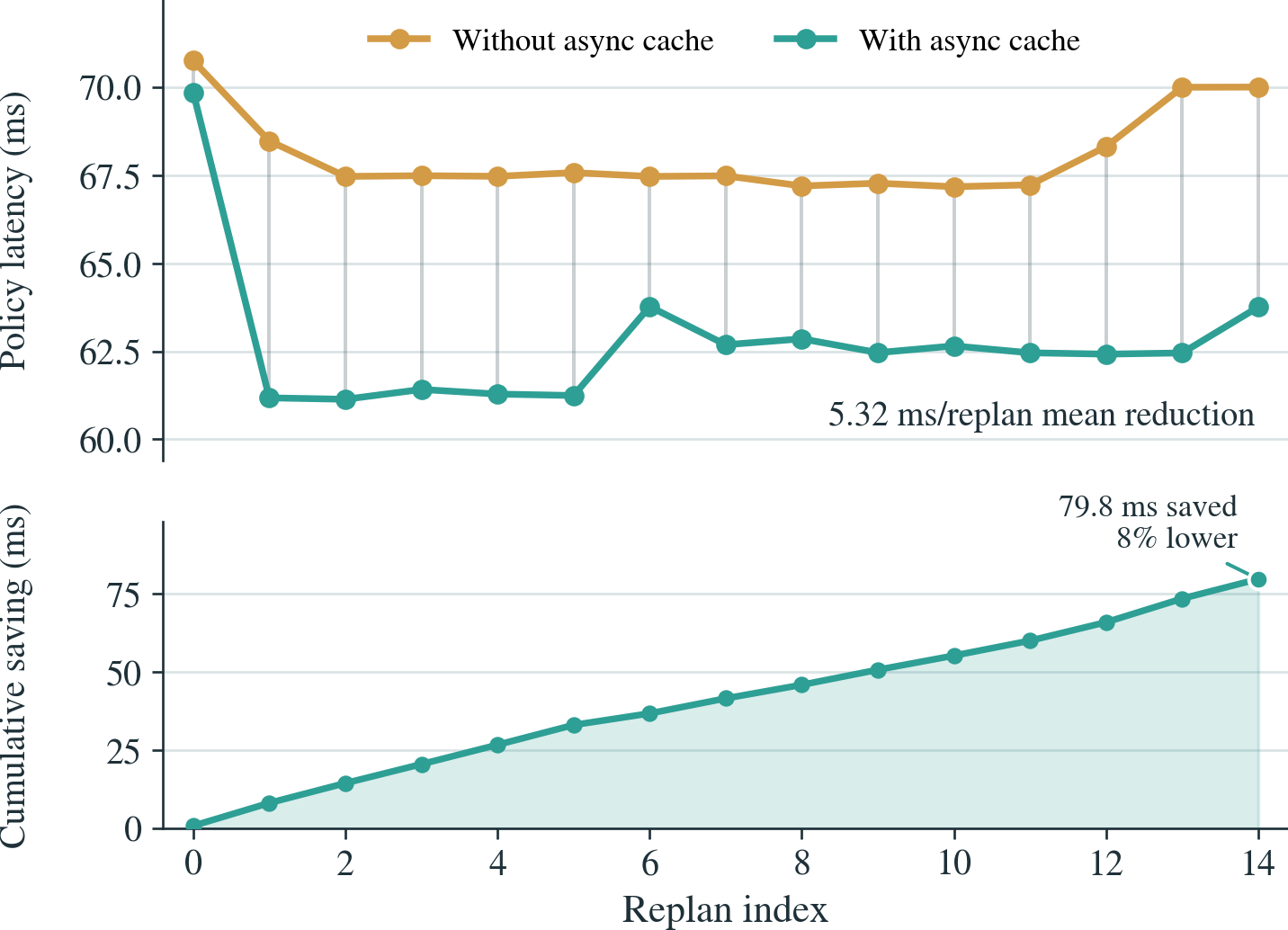}
    \caption{Asynchronous historical-feature caching on LIBERO Long. Top: per-replan server-side policy sampling latency with and without caching. Bottom: cumulative latency saved over the same 15-replan segment.}
    \label{fig:cache_latency}
\end{figure}

As shown in Fig.~\ref{fig:cache_latency}, caching reduces mean latency from $68.10$ to $62.78\,\mathrm{ms}$, corresponding to $5.32\,\mathrm{ms}$ saved per replan. Over the measured segment, cumulative sampling time decreases from $1.021$ to $0.942\,\mathrm{s}$, a $79.8\,\mathrm{ms}$ ($7.8\%$) reduction. At episode start, the asynchronous cache is bypassed until both required historical tokens are available; subsequent replans reuse cached history, removing redundant encoding while current-observation encoding and action sampling remain on the foreground path.

\subsection{Real-Robot Evaluation}

\begin{figure*}[!t]
    \centering
    \includegraphics[width=0.69\textwidth]{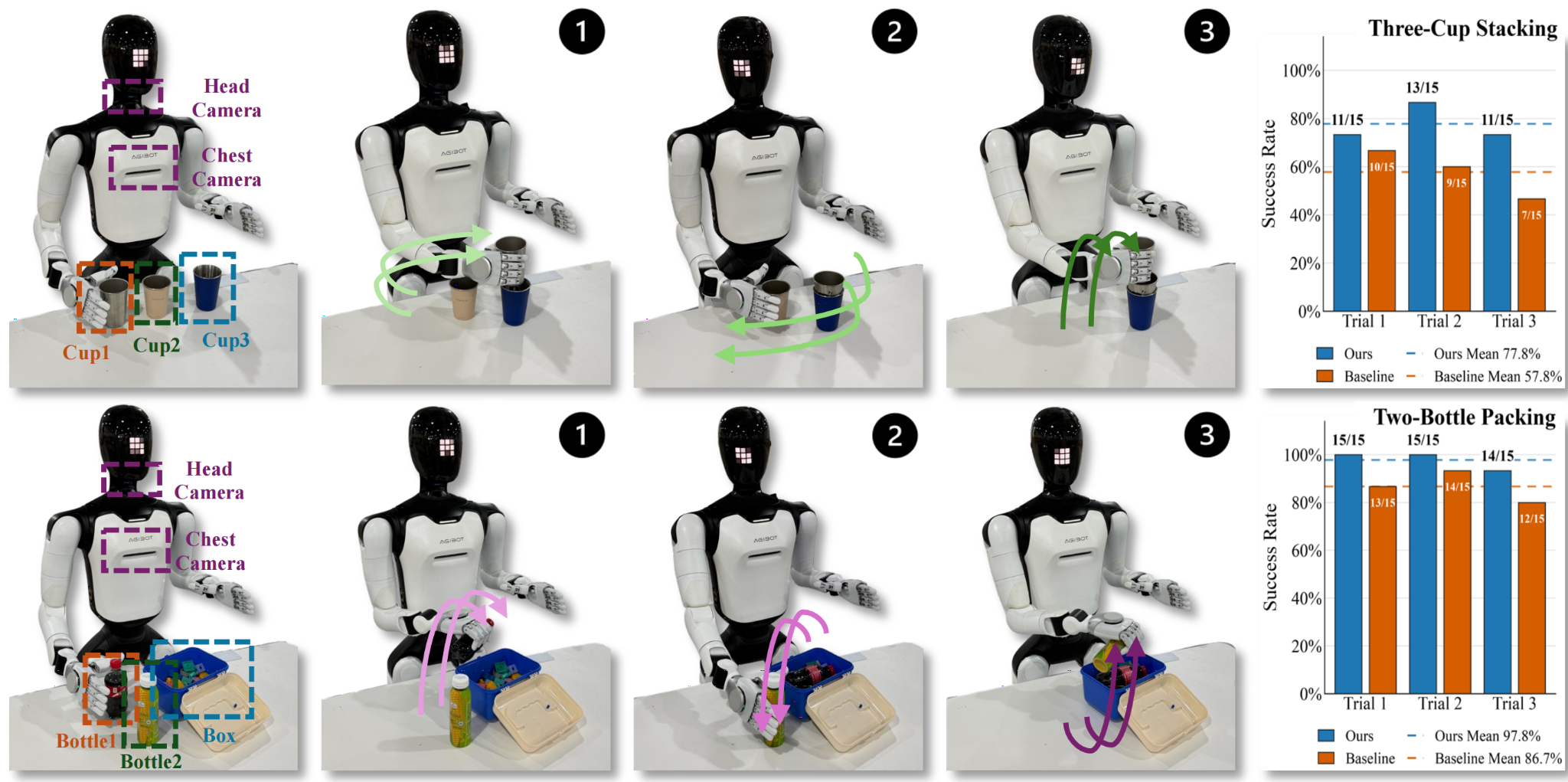}
    \caption{Real-robot evaluation on the AgiBot A3. Left: head/chest views and task setups; the head view provides training-only temporal-flow supervision. Middle: representative three-stage executions. Right: Ours and Baseline over three 15-trial rounds; dashed lines show the 45-trial means.}
    \label{fig:real_robot}
\end{figure*}

We evaluate physical transfer on an AgiBot A3 using two three-stage manipulation tasks, \emph{Three-Cup Stacking} and \emph{Two-Bottle Packing} (Fig.~\ref{fig:real_robot}). Both policies receive head- and chest-view RGB observations, while only the head view is used to construct the training-time temporal-flow supervision. Each task contains 280 demonstrations, and the policies are trained for 60k steps with the common 8-H100, per-GPU batch-size-32 setting. For evaluation, we conduct three evaluation rounds of 15 trials for each task, giving 45 trials per method and task.

TemporalFlow-VLA consistently improves over the baseline in the physical setting. On \emph{Three-Cup Stacking}, mean success increases from $57.8\%$ to $77.8\%$ (+20.0 points); on \emph{Two-Bottle Packing}, it rises from $86.7\%$ to $97.8\%$ (+11.1 points). The gain is larger on cup stacking, where success requires preserving progress across several sequential placements and alignment steps. Together with the simulation results, these experiments indicate that the learned temporal representation remains useful when observations and executions are subject to real-world variation, without requiring geometric inputs at deployment.

\section{CONCLUSION}
We presented TemporalFlow-VLA, a history-aware VLA that learns compact execution history from physically grounded temporal supervision. Instead of treating past observations as additional visual context, we construct robot-surface temporal flow from recorded joint states, robot geometry, and calibrated cameras, and use it to supervise two execution-aligned temporal queries that provide structured history to the action expert. Experiments show that TemporalFlow-VLA achieves $97.63\pm0.26\%$ average success on LIBERO and $85.5\%/84.2\%$ Clean/Randomized success across 12 challenging RoboTwin tasks, with the clearest gains on multi-stage manipulation. Controlled history interventions further show that action prediction depends on both historical content and its correct temporal order, while direct multi-frame conditioning remains consistently weaker than the proposed representation. Finally, asynchronous historical-feature caching reduces server-side policy sampling time by $7.8\%$ over a matched multi-replan control segment. Together, these results suggest that explicitly supervising how recent robot motion manifests in visual observations provides a more effective and deployment-efficient way to incorporate execution history into VLA policies.

An important direction for future work is to more systematically investigate the temporal scale of history used by VLA policies. In this work, we adopt a fixed set of historical observations rather than exhaustively studying how the number and temporal spacing of historical frames affect temporal representation learning. Exploring the optimal temporal horizon and sampling granularity for different manipulation tasks may further improve the effectiveness and generality of temporal information extraction in VLA models.

\section*{ACKNOWLEDGMENT}
OpenAI ChatGPT was used during manuscript preparation to assist with language revision, the refinement of selected textual and visual presentation elements, LaTeX figure and table placement, and code completion within author-developed implementations. All technical decisions and contributions, including the methodology, implementation logic, experimental design, evaluation, interpretation of results, and conclusions, were developed, reviewed, and verified by the authors.

\balance
\bibliographystyle{IEEEtran}
\bibliography{references}

\end{document}